\documentclass{article} 
\usepackage{iclr2026_conference,times}

\usepackage{amsmath,amsfonts,bm}

\def\eqref#1{equation~\ref{#1}}

\def\1{\bm{1}}

\DeclareMathAlphabet{\mathsfit}{\encodingdefault}{\sfdefault}{m}{sl}
\SetMathAlphabet{\mathsfit}{bold}{\encodingdefault}{\sfdefault}{bx}{n}

\usepackage[utf8]{inputenc} 
\usepackage[T1]{fontenc}    
\usepackage{hyperref}       
\usepackage{url}            
\usepackage{booktabs}       
\usepackage{amsfonts}       
\usepackage{nicefrac}       
\usepackage{microtype}      
\usepackage{xcolor}         
\usepackage{amsmath}
\usepackage{mathtools}
\usepackage{amsthm}
\usepackage{graphicx}
\usepackage{subcaption}
\usepackage{multirow}
\usepackage{tabularx}
\usepackage{adjustbox}
\usepackage{wrapfig}
\usepackage{placeins}
\usepackage{natbib} 
\usepackage{enumitem}

\usepackage[table,xcdraw]{xcolor}

\usepackage[linesnumbered,ruled,vlined]{algorithm2e}
\usepackage{amsmath}  
\SetKwInOut{KwIn}{Input}
\SetKwInOut{KwOut}{Output}

\setlist[enumerate]{itemsep=1pt, parsep=0pt}

\newcommand{\tablestyle}[2]{\def\arraystretch{#2}\setlength{\tabcolsep}{#1}}
\newlength\savewidth\newcommand\shline{\noalign{\global\savewidth\arrayrulewidth
  \global\arrayrulewidth 1pt}\hline\noalign{\global\arrayrulewidth\savewidth}}
\usepackage{hyperref}
\usepackage{url}

\newcommand{\taxiemoji}{\includegraphics[height=1em]{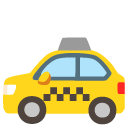}}

\iclrfinalcopy

\title{\taxiemoji\ Data Efficient Any Transformer-to-Mamba Distillation via Attention Bridge}

\author{
\quad\quad\quad\;\;\;\;\;\;\;\textbf{Penghao Wang}$^{1}$\quad 
\textbf{Yuhao Zhou}$^{1}$\footnotemark[4]\quad 
\textbf{Mengxuan W}u$^{1}$\quad 
\textbf{Panpan Zhang}$^{1}$\quad \\ 
\quad\quad\quad\quad\quad\quad\quad\quad\quad\quad\quad\quad\textbf{Zhangyang Wang}$^{2}$\quad 
\textbf{Kai Wang}$^{1}$\footnotemark[2]\quad \\
\quad\quad\quad\quad\quad$^{1}$National University of Singapore \quad $^{2}$The University of Texas at Austin
}

\begin{document}

\maketitle
\begin{abstract}
State-space models (SSMs) have emerged as promising alternatives to Transformers for sequence
modeling. However, training competitive SSMs from scratch remains computationally intensive, and the
ecosystem around them is far less mature than that of Transformers. Moreover, the architectural
differences between SSMs and Transformers make it challenging to efficiently transfer knowledge from
pretrained Transformers. In this work, we propose \textbf{C}ross-architecture distillation via
\textbf{A}ttention \textbf{B}ridge (\textbf{CAB}), a distillation framework that transfers
attention-related representations from Transformer teachers to state-space student models. Unlike
conventional knowledge distillation that supervises only final predictions, CAB enables token-level
intermediate supervision through a lightweight bridge and flexible layer-wise alignment. By aligning
Transformer attention-related representations with Mamba's token-dependent state projections, CAB
facilitates efficient cross-architecture knowledge transfer without inference-time overhead.
Experiments on image classification and language modeling demonstrate that CAB improves
Transformer-to-SSM distillation, particularly under limited-supervision settings. Overall, CAB
provides an efficient pathway for transferring Transformer representations to SSM-based models,
helping bridge the gap between mature Transformer ecosystems and emerging SSM ecosystems.
Our project is available at \href{https://github.com/wph6/CAB}{https://github.com/wph6/CAB}.

\end{abstract}

\section{Introduction}\label{sec:intro}

\begin{figure*}[t]
    \centering

    \begin{minipage}{\linewidth}
        \centering
        \begin{subfigure}[b]{\linewidth}
            \centering
            \includegraphics[width=\linewidth]{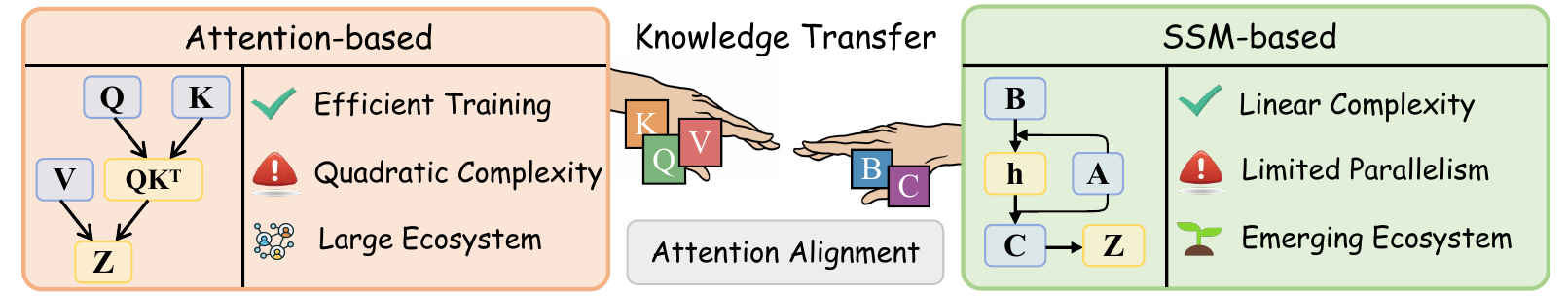}
            \caption{Bridging Attention-based and SSM-based Models via Knowledge Transfer.}
            \label{fig:top}
        \end{subfigure}
    \end{minipage}

    \vspace{4mm}

    \begin{minipage}{\linewidth}
        \centering
        \begin{subfigure}[b]{\linewidth}
            \centering
            \includegraphics[width=\linewidth]{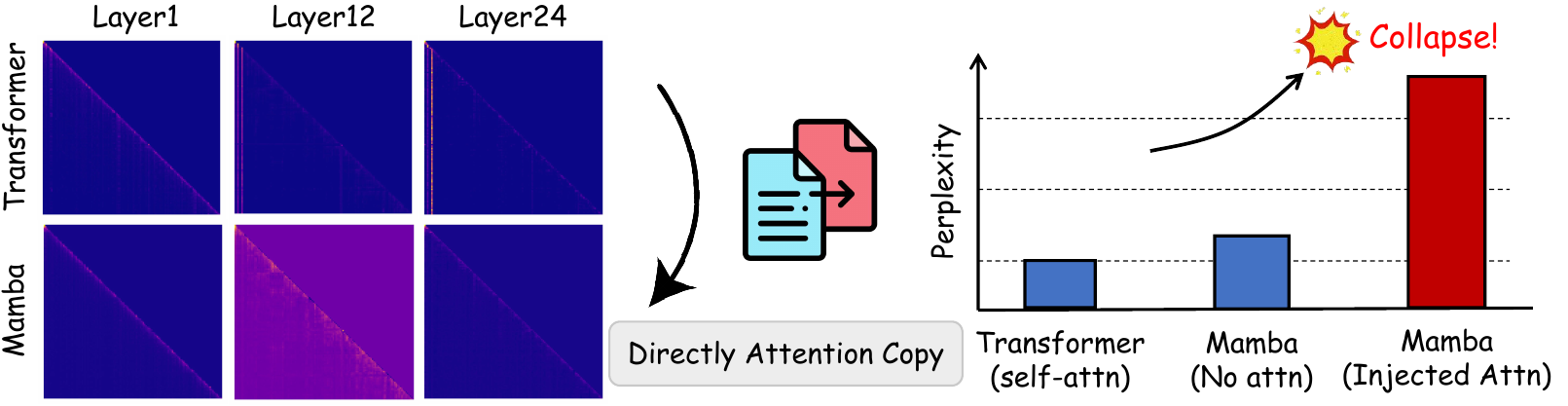}
            \vspace{0.1pt}
            \caption{Left: Attention maps in Transformer and Mamba show partial structural similarity. 
            Right: Directly injecting Transformer attention into Mamba causes performance collapse.}
            \label{fig:bottom}
        \end{subfigure}
    \end{minipage}
    \caption{\textbf{Towards Effective Attention-to-SSM Distillation.}
    We highlight the structural complementarity between attention-based and SSM-based models, and
    the limitations of direct attention transfer, motivating our proposed alignment-based
    distillation approach.}
    \label{fig:intro}
\end{figure*}

Linear RNNs (Mamba \citep{gu2024mamba1}, RWKV \citep{peng2023rwkv}) have re-emerged as a promising
alternative to attention-based models through recurrent state transitions. In contrast, Transformers
use explicit attention to model long-range interactions with high expressiveness, but at the cost of
quadratic computation and limited efficiency on long sequences
\citep{katharopoulos2020transformersarernn}. Among linear RNNs, Mamba exhibits strong long-range
modeling with excellent runtime efficiency \citep{mamba2}.
However, training competitive Mamba models from scratch at scale remains computationally demanding,
and the associated tooling, pretrained checkpoints, and training recipes are less standardized than
those of Transformers. This highlights a challenge: \textit{how can we cost-effectively transfer the
rich inductive biases and knowledge of pretrained Transformers into Mamba architectures}
(Fig.~\ref{fig:top})?


A straightforward approach is knowledge distillation (KD), where a pretrained Transformer guides the
Mamba student. However, this strategy has several limitations: (1) it lacks explicit transfer of
attention-related inductive bias, weakening long-range dependency modeling
\citep{li2024attentransfer,wang2020minilm}; (2) it yields weak gradient signals due to long
backpropagation paths from the output layer, limiting learning in deep architectures
\citep{romero2014fitnets,sun2019patient}; and (3) it ignores architectural differences, leading to
suboptimal knowledge transfer \citep{lu2022bridging,abnar2020transferring}. Although SSMs lack
explicit token interactions, their internal representations still exhibit structural similarity to
Transformer attention maps (Fig.~\ref{fig:bottom}, left), motivating aligned transfer strategies.
However, as shown in Fig.~\ref{fig:bottom}, right, directly injecting Transformer attention into
Mamba leads to performance collapse, underscoring the need for structure-aware knowledge transfer.

\begin{figure*}[t]
    \centering
    \begin{subfigure}[b]{0.32\textwidth}
        \centering
        \includegraphics[width=\textwidth]{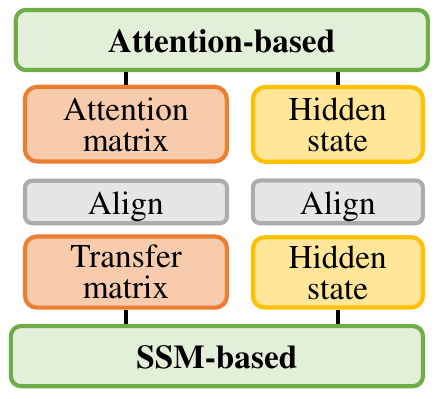}
        \caption{Multi-stage alignment}
        \label{fig:alignment}
    \end{subfigure}
    \hfill
    \begin{subfigure}[b]{0.32\textwidth}
        \centering
        \includegraphics[width=\textwidth]{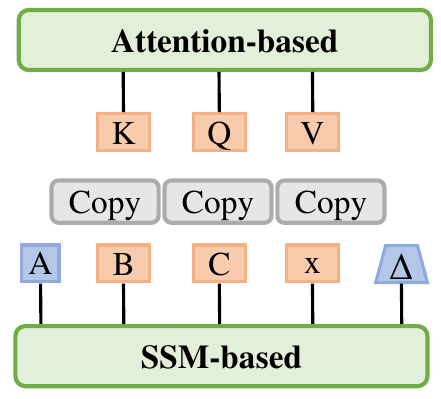}
        \caption{Attention Weight Reuse}
        \label{fig:weight_copy}
    \end{subfigure}
    \hfill
    \begin{subfigure}[b]{0.32\textwidth}
        \centering
        \includegraphics[width=\textwidth]{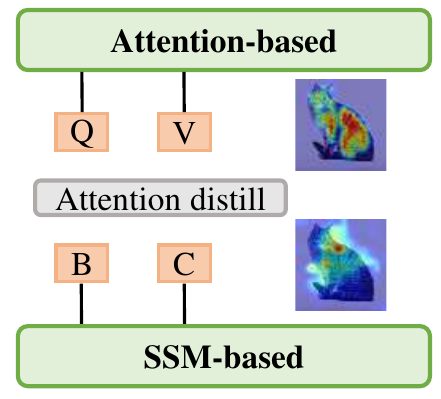}
        \caption{CAB(Ours)}
        \label{fig:attention_distill}
    \end{subfigure}
    
    \caption{Overview of representative strategies for transferring attention-based inductive
    biases from Transformers to State Space Models (SSMs).}
    \label{fig:motivation}
\end{figure*}

Recent work on bridging attention-based and SSM-based architectures still faces key limitations.
MOHAWK \citep{bick2024transformers2ssm} introduce a principled multi-stage alignment of full
attention matrices and hidden states across architectures (Fig.~\ref{fig:alignment}). Their high
memory consumption and training cost make it impractical to scale this attention alignment to large
training volumes \citep{bick2025llamba}. Wang et al. \citep{wang2024mambainllama} attempts to
connect Transformers and Mamba by reusing projection weights from Transformer
layers(Fig.~\ref{fig:weight_copy}), but direct substitution ignores input-dependent dynamics. Both
methods rely on strict layer-to-layer correspondence between the teacher and the student. In
practical settings, the teacher and the student often differ substantially in model depth and scale.

Furthermore, existing approaches often rely on expensive intermediate alignment
procedures, which introduce substantial memory and computational overhead and
limit their scalability to large-scale cross-architecture distillation.
Attention-based Transformers are notoriously data-hungry, requiring large amounts of data to learn
robust token relationships and transferable inductive biases~\citep{kaplan2020scaling,ViT,
zhai2022scaling}. Given such a pretrained Transformer teacher, it becomes desirable to transfer its
learned attention priors to an SSM student under limited labeled supervision. This motivates our
student-side data-efficient distillation setting, where CAB enables Mamba to inherit Transformer
attention knowledge under limited labeled supervision.

To address these limitations, we propose \textbf{CAB}, a novel framework for transferring
attention-based inductive biases into Mamba models ( Fig.~\ref{fig:attention_distill}). We introduce
a lightweight MLP-based \textbf{bridge} between Transformer and Mamba architectures, enabling
efficient fine-grained supervision of attention. Furthermore, we introduce a hierarchical mapping
strategy to bridge the architectural gap between Transformers and Mamba, allowing for more effective
cross-architecture knowledge transfer. Experiments on image classification demonstrate strong
limited-supervision performance, while language modeling results provide supporting evidence that
CAB transfers beyond vision architectures. Full-data ImageNet experiments further reveal when
cross-architecture alignment helps and when prolonged alignment becomes overly restrictive
(Appendix Fig.~\ref{fig:train_curves}).

Our contributions are summarized as follows:

\begin{itemize}

    \item We introduce \textbf{CAB}, a novel distillation framework with a lightweight MLP-based
    \textbf{Attention Bridge} that directly connects the internal representations of teacher and
    student models. This design enables fine-grained, token-level supervision and facilitates
    knowledge transfer across heterogeneous architectures.

    \item CAB avoids explicitly constructing dense attention matrices and instead performs
    token-wise projection alignment, substantially reducing the memory and computational overhead
    of intermediate representation distillation. It is particularly effective when the student is
    trained with a limited number of unique training samples.

    \item Experiments on image classification and language modeling demonstrate the effectiveness
    of CAB across different model scales and architectural configurations. CAB yields particularly
    strong improvements in low-data ImageNet settings and consistently outperforms representative
    Transformer-to-SSM distillation baselines under matched training budgets.
    
\end{itemize}

\section{Related Work}
\paragraph{Linear Sequence Models.}Recent research has explored alternatives to traditional
attention mechanisms for efficient sequence modeling. Linear attention mechanisms retain the
query-key-value structure of Transformers but replace softmax with activation function mappings to
achieve linear time complexity. Efficient Attention \citep{shen2021efficient} was among the first to
decouple the attention map computation into two linear projections, reducing time and memory costs.
Performer \citep{Performer} approximates softmax attention using orthogonal random features with
theoretical guarantees under linear complexity. CosFormer \citep{qincosformer} and Linear
transformers \citep{katharopoulos2020transformersarernn} both adopt feature-mapped formulations of
attention, introducing reweighting or recurrence structures that support causal and efficient
computation.

In parallel, another line of work revisits linear recurrence as an alternative mechanism for
sequence modeling, leading to renewed interest in linear RNNs and state space models. S4 \citep{S4}
introduced structured state matrices for efficient, parallel, and long-range sequence modeling.
Mamba \citep{gu2024mamba1,zeng2025mambaic,li2026mambaraw} extended this with input-dependent,
token-wise state transitions. Mamba-2 \citep{mamba2} further improves speed and scalability through
structured state-space duality (SSD) and hardware-efficient algorithm design. RWKVs
\citep{peng2023rwkv,peng2024rwkv56,peng2025rwkv7} is a recent line of recurrent models that utilize
Time-mix for temporal dependency modeling and Channel-mix for feature transformation.

\paragraph{Knowledge Distillation and Attention Transfer.} Knowledge distillation (KD) aims to
transfer learned knowledge from a large teacher model to a smaller student model. Originally
proposed by Hinton et al. \citep{hinton2015distilling} through logit-based distillation, KD has
since been successfully applied across a variety of domains, including image classification
\citep{romero2014fitnets}and language understanding \citep{sun2019patient}. Within the Transformer
framework, recent work has explored richer supervision signals beyond output logits. TinyBERT
\citep{jiao2020tinybert} and MiniLM \citep{wang2020minilm} distill attention distributions and
intermediate hidden states, showing that aligning multi-level information improves downstream
performance. Other works \citep{zagoruyko2017paying,li2024attentransfer} highlight the
transferability of attention maps in both language and vision domains, suggesting that attention
matrices encapsulate inductive biases useful for generalization.

However, most existing attention distillation methods assume that the teacher and student share
similar Transformer architectures. In contrast, our work addresses the more challenging setting of
cross-architecture distillation from Transformer teachers to state-space model students.

\section{Method}\label{sec:method}
\subsection{Preliminary}
The standard attention mechanism models token interactions using a global attention matrix:
\begin{equation}
\text{Attention}(Q,K,V)=\text{softmax}\left(QK^\top\right)V,
\end{equation}
where $Q, K, V$ represent the query, key, and value projections, respectively. While this
formulation enables powerful context modeling, its quadratic time and space complexity with respect
to sequence length $L$ significantly limits scalability in long-sequence tasks.

To address the inefficiency of standard attention, linear attention was proposed by approximating
the softmax-based attention weights using activation function mappings:
\begin{equation}
\text{Attention}(Q,K,V) = \phi(Q)\varphi(K)^\top V,
\end{equation}
where $\phi(\cdot)$ and $\varphi(\cdot)$ denote activation functions. Under causal constraints, this
formulation admits a recursive structure:
\begin{equation}
h_t = h_{t-1} + \varphi(k_t)^\top v_t, \quad
y_t = \phi(q_t) h_t,
\end{equation}
where $h_t$ acts as a memory of past key--value information and $\phi(q_t)$ performs a dynamic
readout. This reduces complexity to $\mathcal{O}(L)$ and makes linear attention structurally close
to a recurrent model.

Mamba is a state-space model that encodes token interactions implicitly via dynamic recurrence:
\begin{equation}
y_t=C_t\sum_{j=1}^t\left(\Pi_{k=j+1}^t\bar{A}_k\right)\bar{B}_jx_j.
\end{equation}
where the output $y_t$ aggregates all past inputs $\{x_j\}_{j=1}^t$ modulated by the transition
matrices $(\bar{A}, \bar{B}_t, C_t)$. When the transition matrix $\bar{A}$ approximates the
identity, this recurrence simplifies into a form structurally equivalent to the recursive
formulation of linear attention:

\begin{equation}
\begin{aligned}
\left\{
\begin{aligned}
 & h_{t} = h_{t-1} + \varphi(k_t)^\top v_t, \\
 & y_{t} = \phi(q_t) h_t
\end{aligned}
\right.
\Longleftrightarrow\;
\left\{
\begin{aligned}
    h_t &= \bar{A}_t h_{t-1} + \bar{B}_t x_t, \\
    y_t &= C_t h_t
\end{aligned}
\right.
\end{aligned}
\label{eq:linear}
\end{equation}

This perspective provides a principled bridge between attention and state-space models: under this
approximation, aligning the token-dependent projections $B$ and $C$ with the Transformer's $K$ and
$Q$ becomes a natural and theoretically motivated strategy. Prior work further shows that $Q,K$
encode sufficient inductive biases for transferring attention
\citep{li2024attentransfer,clark2019does}. We therefore regard $B_t,C_t$ as \textbf{implicit
attention carriers} and align them with the teacher's $Q,K$, providing fine-grained supervision
without incurring quadratic overhead from dense attention maps.

\subsection{Attention Bridge}  
Inspired by recent representation alignment methods REPA \citep{yurepresentation}, we propose a
lightweight \emph{MLP-based bridge} that links the attention mechanisms of Transformer and Mamba.
Specifically, the bridge maps $(Q,K)$ from the Transformer's explicit attention to $(B,C)$ in
Mamba's implicit attention, adapting to input-dependent dynamics. This design enables
\emph{fine-grained, token-level supervision}: each token's role in the attention space is aligned
with its counterpart in the state-space recurrence.

\paragraph{Attention Alignment.} Existing attention distillation methods
\citep{zagoruyko2017paying,jiao2020tinybert} typically rely on explicit attention matrices as
supervision targets. However, in state-space models like Mamba, attention is implicit and cannot be
directly extracted without prohibitive quadratic overhead, as in full attention alignment methods
\citep{bick2024transformers2ssm}. This gap motivates the need for a new distillation paradigm that
leverages internal token-wise projections rather than explicit pairwise interactions. While
$\bar{B}\!\leftrightarrow\!K$ and $C\!\leftrightarrow\!Q$ provide a bridge in the \emph{discretized}
SSM, we align the \emph{continuous-time}, token-dependent $B$ and $C$ instead, to retain
input-specific dynamics.

However, the representation spaces of $B, C \in \mathbb{R}^{L \times d_s}$ and the teacher's
attention vectors $K, Q \in \mathbb{R}^{L \times d_t}$ are not inherently aligned---neither in
\textbf{dimensionality }nor in \textbf{semantics}. To address this mismatch, we introduce learnable
MLP-based projection modules that map the student's $B$ and $C$ into the teacher's attention space.
These modules implicitly learn activation mappings and dimensional adaptation, enabling a flexible
and effective form of attention-level knowledge transfer across heterogeneous architectures. The
attention alignment loss is then defined as:
\begin{equation}
\begin{split}
\mathcal{L}_{\text{attn}} = \frac{1}{L} \sum_{l=1}^{L} \Bigl(
&\left\| \phi_B(B^{(l)}) - K^{(l)} \right\|_2^2 \\
&+ \left\| \phi_C(C^{(l)}) - Q^{(l)} \right\|_2^2
\Bigr).
\end{split}
\end{equation}


where $l$ indexes student layers and $\phi_B$, $\phi_C$ are MLPs. These projection modules are
introduced only during training and are fully discarded after distillation, incurring no additional
parameters or inference-time computation in the final student model. This alignment avoids explicit
attention maps and supports distillation under limited supervision, making it suitable for
student-side limited-supervision scenarios.

\paragraph{Layer-wise Alignment across Architectures.} A key challenge in cross-architecture
distillation lies in the structural mismatch between attention-based models and SSMs.
Attention-based teachers often contain deep stacks of self-attention layers,  while student models
based on state-space architectures may have a different number of layers due to design choices. This
mismatch in depth makes strict one-to-one layer supervision infeasible and potentially suboptimal.

To address this, we adopt a flexible alignment strategy that allows cross-layer matching. Instead of
enforcing strict one-to-one correspondence, we define a general layer mapping function $g(l)$ that
aligns each student layer $l \in [1, L]$ to a teacher layer via proportional indexing:
\begin{equation}
g(l) = \left\lfloor \frac{l}{L} \cdot T \right\rfloor.
\end{equation}
This unified and relaxed alignment strategy supports both deeper and shallower student networks,
enabling effective knowledge transfer across varying depths and architectural module types while
avoiding over-constraining the student. Moreover, it enhances the transferability of the method
across heterogeneous architectures. The resulting attention alignment loss is:


\begin{equation}
\begin{split}
\mathcal{L}_{\text{attn}} = \frac{1}{L} \sum_{l=1}^{L} \Bigl(
&\left\| \phi_B(B^{(l)}) - K^{(g(l))} \right\|_2^2 \\
&+ \left\| \phi_C(C^{(l)}) - Q^{(g(l))} \right\|_2^2
\Bigr).
\end{split}
\label{eq:attn_loss}
 \end{equation}

\tablestyle{10pt}{1.05}
\begin{table*}[ht]
\centering
\caption{Top-1 accuracy comparison between pretraining and distillation methods on ImageNet
classification under varying proportions of training data.}
\begin{tabular}{ccccccc}
\toprule
Teacher & Student & Method & 1\% & 5\% & 10\% & 20\%  \\
\midrule
- & Vim-Tiny & Vanilla Training & 11.2 & 27.4 & 32.9 & 41.5  \\
\midrule
\multirow{4}{*}{DeiT-Tiny} & \multirow{4}{*}{Vim-Tiny}
& Soft Distillation & 20.4 & 37.0 & 42.0 & 42.7  \\
& & Attention Reuse & 21.3 & 37.1 & 41.5 & 42.6  \\
& & MOHAWK & 21.5 & 36.8 & 45.1 & 44.0  \\
& & \cellcolor{blue!10}\textbf{CAB}(Ours) 
& \cellcolor{blue!10}\textbf{27.8} 
& \cellcolor{blue!10}\textbf{45.4} 
& \cellcolor{blue!10}\textbf{49.2} 
& \cellcolor{blue!10}\textbf{49.4} \\
\midrule
- & Vim-Small & Vanilla Training & 15.3 & 29.2 & 36.2 & 47.0  \\
\midrule
\multirow{4}{*}{DeiT-Small} & \multirow{4}{*}{Vim-Small}
& Soft Distillation & 23.3 & 41.1 & 49.4 & 54.0  \\
& & Attention Reuse & 23.5 & 41.5 & 49.3 & 54.3  \\
& & MOHAWK & 24.0 & 42.1 & 50.1 & 53.9  \\
& & \cellcolor{blue!10}\textbf{CAB}(Ours) 
& \cellcolor{blue!10}\textbf{27.3} 
& \cellcolor{blue!10}\textbf{46.0} 
& \cellcolor{blue!10}\textbf{54.9} 
& \cellcolor{blue!10}\textbf{60.7} \\
\bottomrule
\end{tabular}
\label{tab:imagenet}
\end{table*}

 \section{Experiment}\label{sec:exp}

Transferring attention-related representations across heterogeneous architectures is challenging,
and different domains expose complementary aspects of this problem. We use vision as the primary
evaluation setting, where Transformer attention explicitly models patch interactions while Mamba
captures token dependencies through recurrent states. The limited-supervision ImageNet setting
provides a challenging environment for evaluating the effectiveness of CAB. We further evaluate
language modeling as a supporting setting to test whether the proposed token-wise projection
alignment extends beyond vision tasks.

\subsection{Setup}
We evaluate our proposed attention distillation framework across both vision and language domains.
All student models are trained from random initialization.


\paragraph{Image Classification.} For image classification, we use ImageNet-1k
\citep{deng2009imagenet} and simulate low-resource settings by training on 1\%, 5\%, 10\%, or 20\%
of the training data, sampled per class for label balance. We evaluate CAB under both scale-matched
and depth/width-mismatched teacher--student settings. Specifically, we use \textbf{DeiT} variants
\citep{touvron2021deit} as Transformer teachers and corresponding \textbf{ViM} variants \citep{vim}
as state-space students. Architectural details are summarized in Table~\ref{tab:model_config_all}.

\paragraph{Language Modeling.} To verify that CAB is not limited to vision architectures, we further
evaluate Transformer-to-Mamba distillation in language modeling. In the main paper, DistilGPT2
\citep{bick2024transformers2ssm} serves as the Transformer teacher, Phi-Mamba
\citep{bick2024transformers2ssm} serves as the student, and distillation is performed on
OpenWebText~\citep{Gokaslan2019OpenWeb}. 
Model specifications are provided in Appendix~\ref{appendix:Exp-setting}.

\subsection{Implementation Details.} 
We implement each projection module $\phi_B$ and $\phi_C$ as a 2-layer MLP with SiLU activation
functions across all experiments, which is both simple and effective. The attention bridge is
training-only and incurs no inference-time overhead.
For image classification, As shown in ~\eqref{eq:linear}, initializing $A \approx 0$ (so that
$\bar{A}\approx I$) simplifies the recurrence into a form structurally equivalent to linear
attention, making it natural to align $B$ and $C$ with the teacher's $K$ and $Q$ through our
attention bridge.

Given the bidirectional nature of Vision Mamba, we apply Eq.~\ref{eq:attn_loss} separately to the
forward and backward directions, and sum the two losses:
\begin{equation}
\mathcal{L}_{\text{attn}}^{\text{vision}}
=
\mathcal{L}_{\text{attn}}^{\text{forward}}
+
\mathcal{L}_{\text{attn}}^{\text{backward}}.
\end{equation}




All ViM models are trained using the AdamW optimizer with a learning rate of 5e-4. The batch size is
64, and training is conducted on a single NVIDIA A100 GPU. Standard ImageNet augmentations,
including random cropping and horizontal flipping, are applied during training.  Detailed
experimental settings are provided in Table~\ref{tab:vision_training_recipe}

For language modeling, consistent with the linear-attention interpretation discussed above,
Phi-Mamba (built on Mamba-2) parameterizes $A_i$ as a scalar, which allows us to directly align $B$
and $C$ without special initialization.

To isolate the contribution of our attention alignment mechanism, training is conducted in two
stages: (i) attention alignment, where the student's token-wise projections are matched to the
teacher's key and query representations (Eq.~\ref{eq:attn_loss}); and (ii) soft distillation,
minimizing KL divergence between teacher and student outputs. Both stages are trained on 8 NVIDIA
A100 GPUs, and detailed hyperparameter settings are provided in Table~\ref{tab:llm_training_recipe}


\paragraph{Baselines.} All baselines are trained from random initialization and follow the exact
same training protocol as our method, including data augmentation, optimizer settings, and training
schedules.We define \emph{vanilla training} as standard supervised learning on the same training
subsets, using only the task loss and without any distillation guidance. We compare our method
against several related distillation strategies:
\begin{itemize}
    \item \textbf{Standard Soft Distillation} \citep{hinton2015distilling}: The student model is
    trained to match the teacher's softened output logits using KL divergence loss, without
    supervision on intermediate representations.

    \item \textbf{Mamba in Llama(Attention Reuse)} \citep{wang2024mambainllama}: This method
    distills large language models (LLMs) into hybrid Transformer-Mamba architectures by reusing
    attention weights. We adapt their strategy to our vision and language settings.

    \item \textbf{MOHAWK} \citep{bick2024transformers2ssm}: This approach distills Transformers into
    state-space models by progressively aligning full attention matrices as well as intermediate
    hidden representations. We reimplement their method and apply it in our cross-architecture
    setting.
\end{itemize}
Implementation and architectural details of all baselines, including training configurations and
model specifications, are provided in Appendix~\ref{appendix:Exp-setting}.

\subsection{Visual Results}

\paragraph{Limited-Supervision ImageNet.} \label{Image Classification.} We report image classification
results using matched depth for the teacher and the student to enable direct comparison with
baseline methods. Table~\ref{tab:imagenet} summarizes Top-1 accuracy on ImageNet for various
teacher-student pairs and data ratios. Our method consistently outperforms the standard soft
distillation and other baselines across these settings. Notably, under the DeiT-Tiny teacher and
Vim-Tiny student setting at 10\% data, it achieves up to a \textbf{+16.3\%} improvement over vanilla
training. We further analyze additional architecture variants and the full-data regime below, where
the behavior of CAB reveals how cross-architecture alignment depends on student capacity and
training schedule.

\begin{table}[t]
\centering
\caption{Top-1 accuracy comparison across student architectures and proportions of training data.}
\label{tab:arch_comparison}
\footnotesize
\setlength{\tabcolsep}{2.5pt}
\renewcommand{\arraystretch}{1.1}
\begin{tabular}{@{}llclcccc@{}}
\toprule
Teacher & Student & Params (M) & Method & 1\% & 5\% & 10\% & 20\% \\
\midrule
\multirow{4}{*}{DeiT-Tiny}
  & \multirow{2}{*}{Vim-Mini$^*$} & \multirow{2}{*}{1.1}
  & Soft Distillation & 5.1 & 17.6 & 30.3 & 30.0 \\
  & & & \cellcolor{blue!10}\textbf{CAB (Ours)}
  & \cellcolor{blue!10}\textbf{6.9}
  & \cellcolor{blue!10}\textbf{18.7}
  & \cellcolor{blue!10}\textbf{31.0}
  & \cellcolor{blue!10}\textbf{32.8} \\
\cmidrule(lr){2-8}
  & \multirow{2}{*}{Vim-Mini} & \multirow{2}{*}{2.1}
  & Soft Distillation & 6.6 & 24.0 & 34.6 & 35.3 \\
  & & & \cellcolor{blue!10}\textbf{CAB (Ours)}
  & \cellcolor{blue!10}\textbf{8.0}
  & \cellcolor{blue!10}\textbf{24.1}
  & \cellcolor{blue!10}\textbf{36.0}
  & \cellcolor{blue!10}\textbf{36.2} \\
\midrule
\multirow{4}{*}{DeiT-Small}
  & \multirow{2}{*}{Vim-Tiny$^*$} & \multirow{2}{*}{3.8}
  & Soft Distillation & 7.7 & 27.9 & 42.3 & 41.5 \\
  & & & \cellcolor{blue!10}\textbf{CAB (Ours)}
  & \cellcolor{blue!10}\textbf{8.1}
  & \cellcolor{blue!10}\textbf{28.1}
  & \cellcolor{blue!10}\textbf{42.8}
  & \cellcolor{blue!10}\textbf{42.5} \\
\cmidrule(lr){2-8}
  & \multirow{2}{*}{Vim-Tiny} & \multirow{2}{*}{7.1}
  & Soft Distillation & 7.8 & 27.6 & 43.9 & 45.0 \\
  & & & \cellcolor{blue!10}\textbf{CAB (Ours)}
  & \cellcolor{blue!10}\textbf{8.7}
  & \cellcolor{blue!10}\textbf{30.7}
  & \cellcolor{blue!10}\textbf{44.9}
  & \cellcolor{blue!10}\textbf{47.2} \\
\bottomrule
\end{tabular}
\end{table}

\paragraph{Robustness to Architectural Mismatch.} CAB is designed for heterogeneous teacher-student pairs
whose depths and widths may differ, so we further evaluate additional Vim variants with different model capacities and depths.
The variants marked with $^*$ have manually adjusted depth configurations to simulate
larger teacher-student architectural mismatches.
As shown in Table~\ref{tab:arch_comparison}, CAB consistently improves over soft
distillation across shallow, narrow, and standard student configurations. These results support the
flexible cross-layer alignment strategy: CAB does not rely on a strict one-to-one architectural
match and remains effective across different student scales.


\paragraph{Convergence Analysis on ImageNet.}\label{sec:convergence}
We further analyze the convergence behavior of different training strategies.
Fig.~\ref{fig:convergence} presents these results. Our method converges significantly faster and
achieves lower test loss throughout training, highlighting its optimization efficiency under limited
supervision.

\begin{wraptable}{r}{0.48\textwidth}
    \centering
    \scriptsize
    \setlength{\tabcolsep}{2pt}
    \renewcommand{\arraystretch}{1.1}
    \begin{tabular}{lcc}
        \toprule
        \textbf{Method} & \textbf{Top-1 Acc $\uparrow$} & \textbf{Top-5 Acc $\uparrow$} \\
        \midrule
        DeiT-Tiny (teacher) & 72.20 & 91.10 \\
        Soft Distillation & 74.56 & 92.46 \\
        CAB (35 epoch) & 74.50 & 92.38 \\
        CAB (50 epoch) & \textbf{74.85} & \textbf{92.56} \\
        CAB (300 epoch) & 71.34 & 90.64 \\
        \bottomrule
    \end{tabular}
    \captionsetup{font=small}
    \caption{Full-data ImageNet alignment scheduling.}
    \label{tab:full_data}
\end{wraptable}

\paragraph{Full-Data ImageNet Training.} Although CAB provides strong gains in
limited-supervision settings, we further study how the alignment signal behaves when the full
ImageNet-1K training set is available. Table~\ref{tab:full_data} reports the full-data alignment
schedule; the epoch in parentheses denotes when CAB alignment is stopped before switching to KL
distillation.

As shown in Table~\ref{tab:full_data}, applying CAB
alignment throughout the entire training process is not optimal: CAB(300epoch) achieves lower final
accuracy than standard soft distillation. However, the training curves in
Appendix Fig.~\ref{fig:train_curves} show that CAB provides stronger guidance during the early optimization
stage and accelerates convergence.

This observation suggests that the effectiveness of cross-architecture alignment depends on its
training schedule. Similar to recent findings on early-stopped representation
alignment~\citep{wang2025repa}, we investigate an early-stopped CAB schedule that applies alignment
only during the initial training phase before switching to conventional logit distillation. This
strategy achieves 74.85\% Top-1 accuracy, slightly improving over standard soft distillation
(74.56\%) while avoiding the degradation caused by persistent alignment.

These results reveal a trade-off in cross-architecture representation alignment: strong alignment
signals are beneficial when the student is still learning transferable structures, but prolonged
alignment may restrict the student from adapting its own architecture-specific representations.

\subsection{Language Modeling Results}
To verify that CAB is not limited to vision architectures, we further evaluate
Transformer-to-Mamba distillation in language modeling. Table~\ref{tab:lm_ppl} reports perplexity
results evaluated under two training regimes corresponding to the second training stage, using 2B
and 4B tokens respectively. CAB consistently achieves lower perplexity across all benchmarks and
scales, outperforming both attention reuse and MOHAWK baselines under the same training budgets.

\tablestyle{6pt}{1.1}
\begin{table}[tb]
\centering
\caption{Perplexity comparison on language modeling benchmarks. DistilGPT2 is used as the teacher
and Phi-Mamba as the student.}
\label{tab:lm_ppl}

\begin{adjustbox}{max width=\linewidth}
\begin{tabular}{c|cc|cc|cc}
\toprule
\multirow{2}{*}{Method}
& \multicolumn{2}{c|}{OpenWebText $\downarrow$}
& \multicolumn{2}{c|}{C4 $\downarrow$}
& \multicolumn{2}{c}{WikiText $\downarrow$} \\
& 2B & 4B & 2B & 4B & 2B & 4B \\
\midrule
\textit{Teacher PPL}
& \multicolumn{2}{c|}{\textit{28.2}}
& \multicolumn{2}{c|}{\textit{39.7}}
& \multicolumn{2}{c}{\textit{52.96}} \\
\midrule
Attention Reuse
& 61.8 & 37.2
& 111.0 & 62.9
& 212.3 & 99.8 \\
MOHAWK
& 58.4 & 31.1
& 105.7 & 51.2
& 199.3 & 77.9 \\
\cellcolor{blue!10}\textbf{CAB (Ours)}
& \cellcolor{blue!10}\textbf{54.4} & \cellcolor{blue!10}\textbf{30.1}
& \cellcolor{blue!10}\textbf{97.3} & \cellcolor{blue!10}\textbf{50.1}
& \cellcolor{blue!10}\textbf{175.0} & \cellcolor{blue!10}\textbf{74.7} \\
\bottomrule
\end{tabular}
\end{adjustbox}
\end{table}

Beyond achieving strong performance on the in-distribution OpenWebText corpus, CAB also generalizes
to out-of-distribution datasets such as C4 and WikiText, which differ in domain and style. We
hypothesize that this improvement is partly due to the intermediate supervision provided by CAB,
which encourages the student to align with teacher representations beyond the final next-token
distribution. Specifically, Transformer Q/K projections capture context-selection patterns, while
Mamba B/C projections control information flow through recurrent states. CAB connects these
representations and provides additional guidance that may generalize better across different text
distributions.

\paragraph{Downstream Evaluation at the 1B Parameter Scale.}
\label{sec:llm-scale}
We further evaluate CAB with 1B-scale language model experiments in larger
Transformer-to-SSM transfer settings. Following the training pipeline of
\citet{bick2024transformers2ssm}, we use a Transformer teacher and an SSM student with a two-stage
schedule: 200M tokens for attention alignment and 2.8B tokens for knowledge distillation. Detailed
implementation settings are provided in Appendix~\ref{app:llm-implementation}. As shown in
Table~\ref{tab:llm_scalability}, CAB improves over representative Transformer-to-SSM distillation
baselines under the matched training setup.

\tablestyle{6pt}{1.1}
\begin{center}
\captionof{table}{Downstream evaluation results at the 1B parameter scale.}
\label{tab:llm_scalability}
\centering
\begin{adjustbox}{max width=\linewidth}
\begin{tabular}{lccccc}
\toprule
\textbf{Model} & \textbf{HellaS.} & \textbf{PIQA} & \textbf{ARC-E} & \textbf{ARC-C} & \textbf{WinoG.} \\
\midrule
Attention Reuse  & 29.42 & 62.79 & 36.07 & 21.84 & 51.22 \\
MOHAWK           & 28.87 & 63.11 & 36.24 & 22.70 & 51.22 \\
\cellcolor{blue!10}\textbf{CAB (Ours)}
& \cellcolor{blue!10}\textbf{31.73}
& \cellcolor{blue!10}\textbf{64.64}
& \cellcolor{blue!10}\textbf{37.75}
& \cellcolor{blue!10}\textbf{23.21}
& \cellcolor{blue!10}\textbf{51.85} \\
\bottomrule
\end{tabular}
\end{adjustbox}
\end{center}

\FloatBarrier
\paragraph{Efficiency.}
\begin{wrapfigure}{r}{0.46\textwidth}
    \centering
    \includegraphics[width=\linewidth]{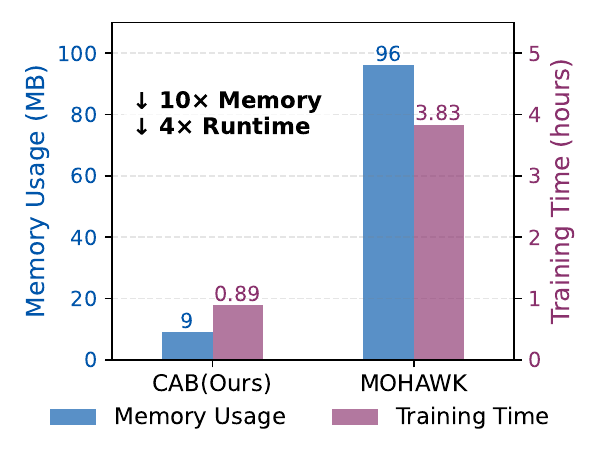}
    \captionsetup{font=small}
    \caption{Alignment-loss memory and runtime comparison of CAB vs.\ full-matrix distillation.}
    \label{fig:memory_time_comparison}
\end{wrapfigure}
CAB reduces the alignment-loss cost of intermediate representation
distillation by avoiding explicit dense attention matrices. At sequence length $L=1024$, aligning
one layer with MOHAWK requires storing $2 \cdot (H,L,L)$ attention tensors for the alignment loss,
whereas CAB aligns token-wise projections with lightweight MLP bridges whose cost depends on model
dimensions rather than the $L^2$ attention matrix. On DistilGPT2 $\rightarrow$
Phi-Mamba-123M with 200M tokens, CAB achieves up to \textbf{$10\times$} lower alignment-loss memory
and \textbf{$4\times$} faster alignment-loss runtime on a single A100 GPU compared with full-matrix
distillation (Fig.~\ref{fig:memory_time_comparison}). These measurements isolate the alignment
objective itself rather than total end-to-end training memory, which also includes parameters,
optimizer states, and activations.

\FloatBarrier
\subsection{Ablation and Representation Analysis}

To evaluate the effectiveness and robustness of our proposed distillation framework, we conduct
ablation studies from three complementary perspectives: (1) the contribution of $B$ and $C$
projection alignment to knowledge transfer, (2) the impact of the transition matrix $\bar{A}$
initialization strategy, and (3) the design choices of the attention bridge, including the
projection architecture and activation functions. All ablation experiments are conducted under the
DeiT-Tiny teacher and Vim-Tiny student setting on a fixed 10\% subset of ImageNet-1k to reduce
computational cost while maintaining task difficulty.

\begin{wraptable}{r}{0.48\textwidth}
    \centering
    \scriptsize
    \setlength{\tabcolsep}{3pt}
    \renewcommand{\arraystretch}{1.05}
    \begin{tabular}{lc}
        \toprule
        Setting & Acc (\%) \\
        \midrule
        Vanilla Training & 32.9 \\
        Soft Distillation & 42.0 \\
        Align $B$ only & 48.7 \\
        Align $C$ only & 48.8 \\
        Shared $\phi$ for $B$ and $C$ & 49.0 \\
        Align $B+C$ w/ default $\bar{A}$ & 43.1 \\
        Align $B+C$ w/ $\bar{A} \approx I$ (Ours) & \textbf{49.2} \\
        \bottomrule
    \end{tabular}
    \caption{Ablation study on distillation strategies.}
    \label{tab:ablation_distill}
\end{wraptable}
\paragraph{Effect of $B$/$C$ Distillation.}\label{sec:ablation_bc_distill} We first investigate
whether aligning both $B$ and $C$ projections is necessary, or if aligning either alone suffices. We
compare three variants: distilling only $B$, only $C$, and both. As shown in
Table~\ref{tab:ablation_distill}, we observe that while distilling either component individually
brings measurable gains over no distillation, jointly aligning $B$ and $C$ yields the best
performance. This supports our hypothesis that $B$ and $C$ jointly encode the implicit attention
mechanism in Mamba, and both are needed to effectively transfer structural knowledge from the
teacher. We also test a variant where $B$ and $C$ share the same projection ($\phi_B \equiv
\phi_C$). As shown in Table~\ref{tab:ablation_distill}, this setting underperforms independent
mappings, since $B$ and $C$ lie in different semantic spaces---$B$ injects inputs while $C$ extracts
context---confirming the need for distinct projections.

\paragraph{Effect of $\bar{A}$ Initialization.}
We study how the initialization of the transition matrix influences the
effectiveness of cross-architecture distillation. In Mamba, $A$ is typically
initialized with a log-spaced diagonal from S4D \citep{gu2022s4d}, which is
designed to capture diverse temporal dynamics in sequence modeling. However,
such initialization may introduce recurrent dynamics that are not well aligned
with attention-based supervision. In our setting, we instead initialize
$A \approx 0$, such that $\bar{A} \approx I$, simplifying the recurrence into
an additive form that resembles linear attention. This provides a more
compatible starting point for aligning Mamba's token-dependent projections with
Transformer representations. As shown in Table~\ref{tab:ablation_distill}, the
attention-compatible initialization substantially improves over the default
S4D initialization, demonstrating that matching recurrent dynamics is important
for effective knowledge transfer across heterogeneous architectures.
Additional ablations on the attention bridge design are provided in Appendix~\ref{sec:bridge_ablation}.
%


\paragraph{Attention Representation Similarity Analysis.}
\label{sec:attention-similarity}
To further understand the underlying attention behavior, we analyze the similarity between the
attention matrices of Vim and a pretrained ViT across all layers
(Fig.~\ref{fig:attention_alignment_comparison}). Before L3, similarity decreases slightly,
reflecting the local focus of early layers \citep{zeiler2014visualizing}. Beyond L3, however,
similarity rises sharply, indicating that our alignment loss effectively guides Vim to mimic
Transformer-style attention. This trend suggests that mid-to-deep layers recover global token
interactions typically lost under standard training, thereby bridging the gap between SSMs and
attention-based models.

\begin{center}
  \centering
  \begin{minipage}[t]{0.48\linewidth}
    \centering
    \includegraphics[width=\linewidth]{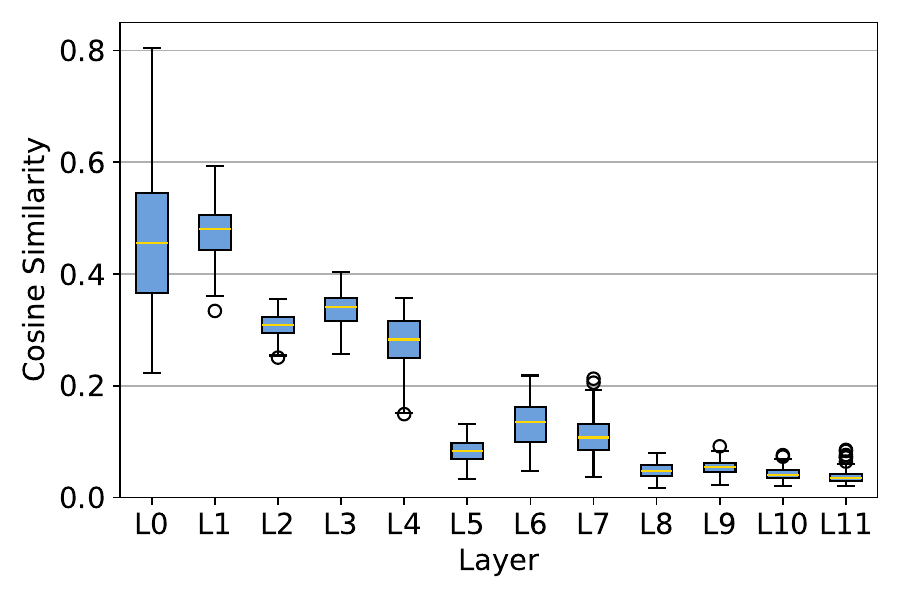}
    \small (a) Vanilla training.
  \end{minipage}%
  \hfill
  \begin{minipage}[t]{0.48\linewidth}
    \centering
    \includegraphics[width=\linewidth]{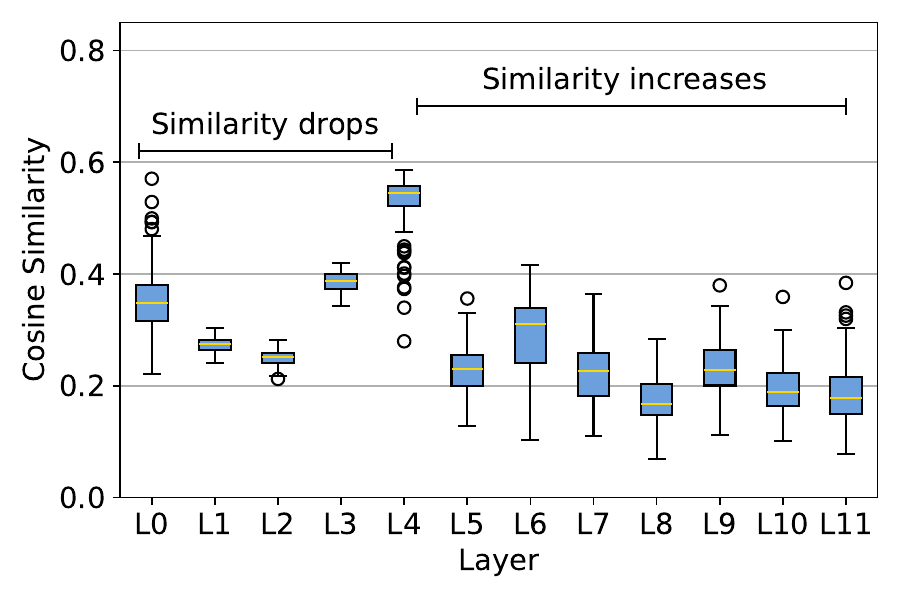}
    \small (b) Attention alignment.
  \end{minipage}
  \captionof{figure}{Cosine similarity between Vim and pretrained ViT, comparing results with and
  without attention alignment. Higher similarity indicates better alignment of attention
  representations.}
  \label{fig:attention_alignment_comparison}
\end{center}

\FloatBarrier
\section{Conclusion}\label{sec:conclu}

In this work, we propose \textbf{CAB}, an attention distillation framework for
transferring attention-related representations from Transformer-based teachers
to state-space student models. CAB introduces an MLP-based
\textbf{Attention Bridge} that aligns Mamba's token-dependent B/C projections
with the teacher's key and query representations, enabling token-level
cross-architecture supervision. Experiments on image classification, together
with language modeling results, demonstrate that CAB improves
Transformer-to-SSM distillation, particularly in student-side
limited-supervision regimes.

These results show that attention-related representations can be effectively
transferred to recurrent models without the overhead of explicit attention
matrix alignment. Our findings highlight the potential of cross-architecture
representation alignment for improving SSM-based models. Extending CAB beyond
Mamba and Transformers to other SSM variants and hybrid architectures remains
an interesting direction for future work.

\bibliography{iclr2026_conference}
\bibliographystyle{iclr2026_conference}
\clearpage
\appendix
\setcounter{secnumdepth}{2}
\section{Additional Experimental Results}
\label{appendix:additional_experiments}

\subsection{Convergence Analysis on ImageNet}
Figure~\ref{fig:convergence} provides the complete training trajectories underlying the
convergence analysis in the main text. CAB reduces test loss more rapidly and remains below the
other training strategies throughout optimization, corroborating its improved optimization
efficiency under limited supervision.
\begin{center}
    \centering
    \includegraphics[width=0.65\linewidth]{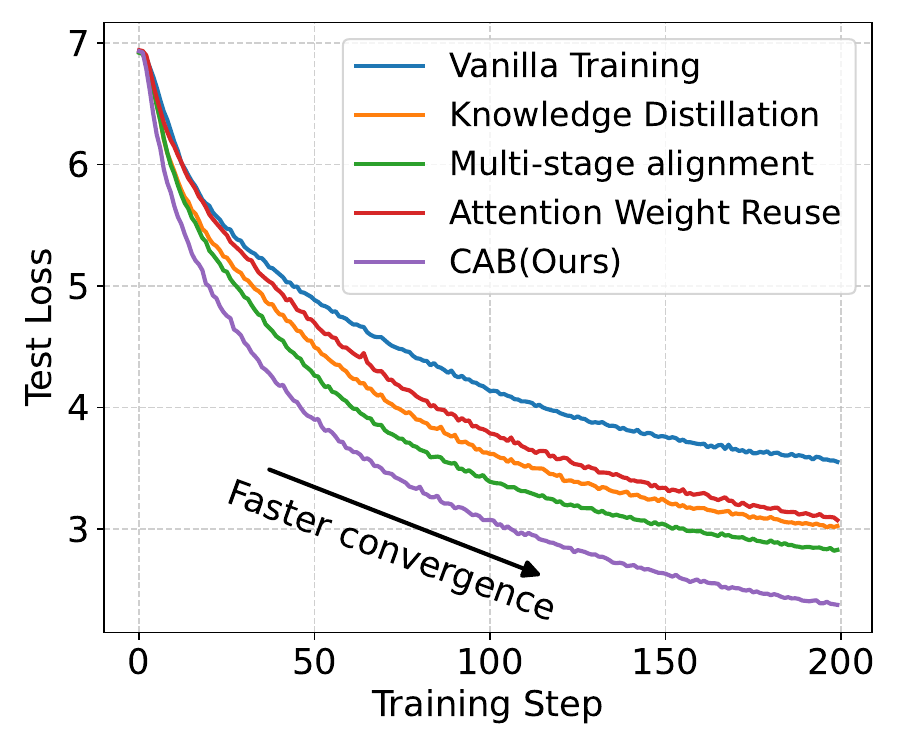}
    \captionof{figure}{Convergence speed and test loss comparison on ImageNet.}
    \label{fig:convergence}
\end{center}

\subsection{Full-Data Alignment Scheduling}
Figure~\ref{fig:train_curves} shows the effect of stopping CAB alignment at different training
epochs. Early-stopped schedules preserve the benefit of alignment, whereas maintaining alignment
for all 300 epochs leads to a lower final accuracy, motivating the schedule used in the main text.
\begin{center}
    \centering
    \includegraphics[width=0.65\linewidth]{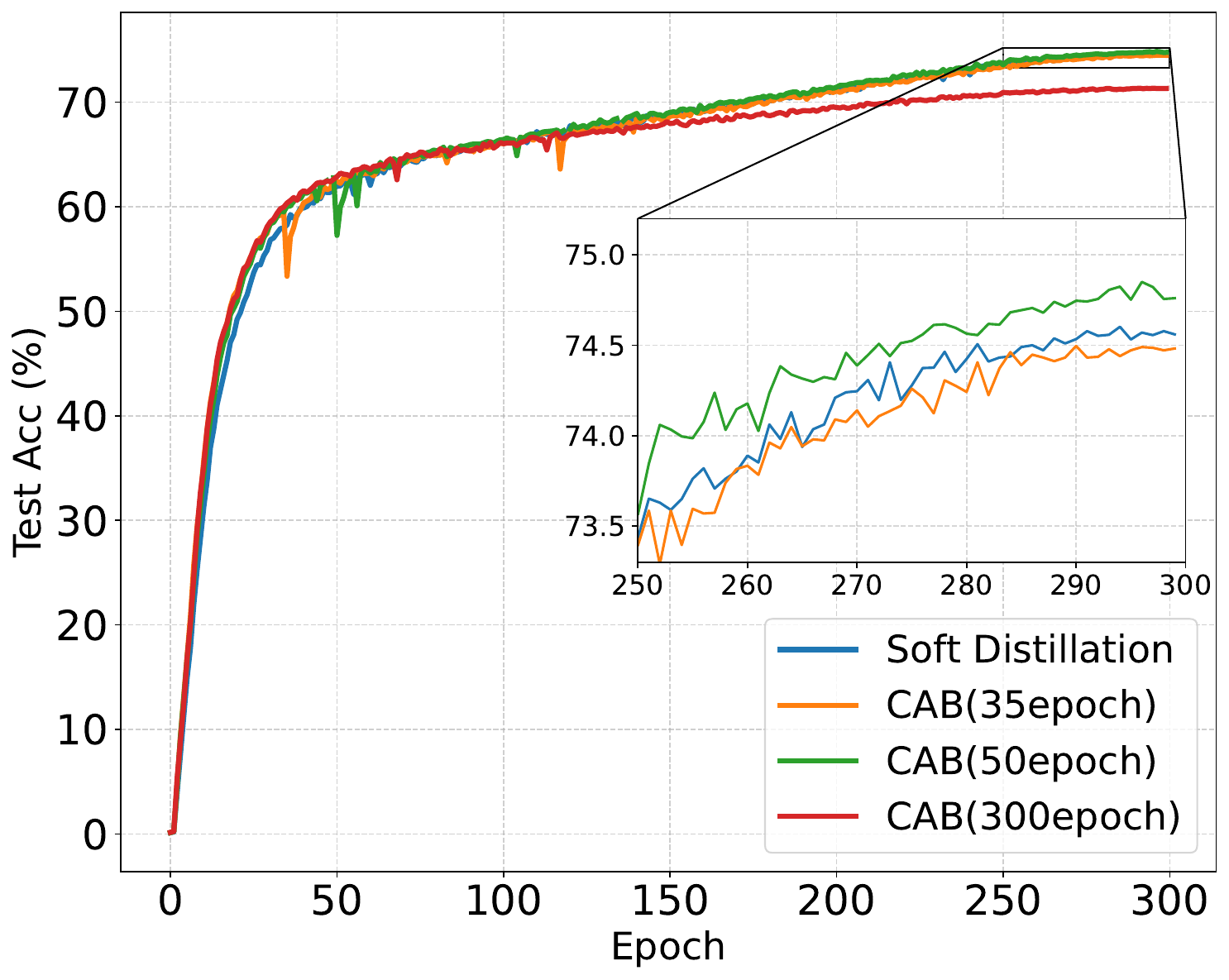}
    \captionof{figure}{Effect of alignment duration on CAB training under full-data ImageNet.}
    \label{fig:train_curves}
\end{center}

\subsection{Full MOHAWK Baseline with Hidden-State Matching}
\label{app:full-mohawk}
Our main language modeling table compares CAB with the two-stage MOHAWK variant under the matched
200M/2B token budget. To address whether this weakens MOHAWK by omitting its hidden-state matching
stage, we further evaluate full three-stage schedules that include attention alignment,
hidden-state matching, and output knowledge distillation. We consider two settings: a fixed-budget
schedule that splits the original 200M intermediate-alignment budget into 50M attention-alignment
tokens and 150M hidden-state-matching tokens, and an extra-budget schedule that keeps the original
200M attention-alignment stage and adds another 200M tokens for hidden-state matching before the 2B
output-distillation stage.

Table~\ref{tab:full_mohawk} shows that CAB remains better than MOHAWK under all corresponding
schedules. Both methods degrade when hidden-state matching is added, suggesting that additional
intermediate matching can introduce conflicting constraints in Transformer-to-SSM distillation
rather than consistently improving transfer.

\tablestyle{2pt}{1.05}
\begin{center}
\centering
\captionof{table}{Language modeling perplexity comparison with full three-stage MOHAWK baselines. Lower is
better.}
\label{tab:full_mohawk}
\begin{adjustbox}{max width=\linewidth}
\begin{tabular}{llccc}
\toprule
\textbf{Method} & \textbf{Stage Schedule} & \textbf{OpenWebText $\downarrow$} &
\textbf{C4 $\downarrow$} & \textbf{WikiText $\downarrow$} \\
\midrule
MOHAWK (2-stage) & 200M / 2B & 58.40 & 105.70 & 199.30 \\
\cellcolor{blue!10}\textbf{CAB (2-stage)} & \cellcolor{blue!10}200M / 2B
& \cellcolor{blue!10}\textbf{54.40}
& \cellcolor{blue!10}\textbf{97.30}
& \cellcolor{blue!10}\textbf{175.00} \\
\midrule
MOHAWK (3-stage) & 50M / 150M / 2B & 73.72 & 133.47 & 271.08 \\
\cellcolor{blue!10}\textbf{CAB (3-stage)} & \cellcolor{blue!10}50M / 150M / 2B
& \cellcolor{blue!10}\textbf{58.62}
& \cellcolor{blue!10}\textbf{105.51}
& \cellcolor{blue!10}\textbf{194.68} \\
\midrule
MOHAWK (3-stage) & 200M / 200M / 2B & 75.55 & 136.33 & 276.82 \\
\cellcolor{blue!10}\textbf{CAB (3-stage)} & \cellcolor{blue!10}200M / 200M / 2B
& \cellcolor{blue!10}\textbf{63.12}
& \cellcolor{blue!10}\textbf{114.55}
& \cellcolor{blue!10}\textbf{217.31} \\
\bottomrule
\end{tabular}
\end{adjustbox}
\end{center}

\FloatBarrier
\section{Additional Analysis and Ablations}

\subsection{Layer Mapping Strategy}
\label{app:layer-mapping}
We further evaluate whether CAB depends on the proportional layer mapping heuristic. Using the same
DeiT-Tiny $\rightarrow$ Vim-Tiny setting with 10\% ImageNet data as in the main ablation study, we
compare proportional matching with two alternatives: random teacher-student layer matching and
matching only to the first half of teacher layers. The bridge design and all training
hyperparameters are kept fixed.

As shown in Table~\ref{tab:layer_mapping}, all three strategies perform similarly, with
proportional matching slightly better. This suggests that the main gain of CAB comes from the
Q/K-to-C/B attention bridge rather than careful tuning of the layer mapping rule.

\tablestyle{6pt}{1.05}
\begin{center}
\centering
\captionof{table}{Ablation study on layer mapping strategies under DeiT-Tiny $\rightarrow$ Vim-Tiny with
10\% ImageNet training data.}
\label{tab:layer_mapping}
\begin{adjustbox}{max width=\linewidth}
\begin{tabular}{llc}
\toprule
\textbf{Mapping Strategy} & \textbf{Description} & \textbf{Top-1 Acc. (\%)} \\
\midrule
Random Matching & Random teacher-student layer pairs & 48.7 \\
First-half Matching & Match to the teacher's early layers & 48.8 \\
\cellcolor{blue!10}\textbf{Proportional Matching (CAB)}
& \cellcolor{blue!10}Match layers by relative depth
& \cellcolor{blue!10}\textbf{49.2} \\
\bottomrule
\end{tabular}
\end{adjustbox}
\end{center}

\subsection{Attention Bridge Design}\label{sec:bridge_ablation}
We ablate the design choices of the attention bridge, including the projection architecture and
activation functions. A linear projection performs substantially worse, indicating that non-linear
mappings are important for cross-architecture alignment. Performance gains saturate beyond two MLP
layers, making a 2-layer MLP a favorable efficiency--accuracy trade-off. Results are also largely
insensitive to the activation choice; we use SiLU by default for consistency across experiments.

\begin{center}
\centering
\captionof{table}{Ablation studies on the attention bridge design.}
\label{tab:bridge_ablation_pair}
\small
\begin{tabular}{l c}
\toprule
\multicolumn{2}{c}{\textbf{Projection architecture}} \\
\midrule
Bridge Variant & Top-1 Acc. (\%) \\
\midrule
Linear projection & 45.0 \\
1-layer MLP & 48.6 \\
2-layer MLP (default) & 49.2 \\
3-layer MLP & \textbf{49.3} \\
\bottomrule
\end{tabular}

\vspace{2mm}

\begin{tabular}{l c}
\toprule
\multicolumn{2}{c}{\textbf{Activation function (2-layer MLP)}} \\
\midrule
Activation & Top-1 Acc. (\%) \\
\midrule
ReLU & 49.1 \\
GELU & \textbf{49.2} \\
SiLU (default) & \textbf{49.2} \\
\bottomrule
\end{tabular}

\end{center}

\FloatBarrier
\section{Experiment Settings}\label{appendix:Exp-setting}
In this section, we detail the experimental settings and provide instructions for reproduction.

\subsection{Model Architectures}
Table~\ref{tab:model_config_all} summarizes the model architectures. Asterisk ($^*$) marks manually
adjusted variants to simulate heterogeneous settings, covering diverse scales across both domains.

\begin{table*}[t]
\centering
\caption{Detailed architectural configurations of teacher and student models used in our experiments.}
\label{tab:model_config_all}
\renewcommand{\arraystretch}{1.2}
\begin{tabular}{lcccccc}
\toprule
\textbf{Model} & $d_{\text{model}}$ & $d_{\text{state}}$ & $d_{\text{conv}}$ & \#Layers & \#Params (M) & FLOPs (G) \\
\midrule
\multicolumn{7}{l}{\textit{\textbf{Vision Models}}} \\
\midrule
Vim-Mini$^*$                          & 96  & 16 & 4  & 12 & 1.1 & 0.15 \\
Vim-Mini                              & 96  & 16 & 4  & 24 & 2.1 & 0.28 \\
Vim-Tiny$^*$                          & 192 & 16 & 4  & 12 & 3.8 & 0.55 \\
Vim-Tiny                              & 192 & 16 & 4  & 24 & 7.1 & 1.08 \\
Vim-Small                             & 384 & 16 & 4  & 24 & 26  & 4.00 \\
DeiT-Tiny                             & 192 & -- & -- & 12 & 5   & 1.26 \\
DeiT-Small                            & 384 & -- & -- & 12 & 22  & 4.61 \\

\midrule
\multicolumn{7}{l}{\textit{\textbf{Language Models}}} \\
\midrule
DistilGPT2                            & 768 & -- & -- & 6  & 88  & 21.68 \\
Phi-Mamba$^*$                       & 768 & 64 & 4  & 6  & 123  & 20.76 \\
Phi1.5          &2048 & -- & -- & 24 & 1418 & 336.11 \\
Phi-Mamba  &2048 & 64 & 4 & 24 & 1521 & 362.30 \\
\bottomrule
\end{tabular}
\end{table*}

\subsection{Implementation Details for Vision Distillation Methods}
To ensure a fair and comprehensive comparison, we extend two representative Transformer-to-SSM
distillation frameworks \citep{bick2024transformers2ssm,wang2024mambainllama}to the vision domain.
While both were originally proposed for language modeling, we adapt them for image classification on
ImageNet-1k by modifying the distillation targets and model-specific components as follows:

\paragraph{Transformers to SSMs (MOHAWK).} The original MOHAWK framework
\citep{bick2024transformers2ssm} consists of three stages: attention matrix alignment, hidden state
alignment, and output-level knowledge distillation. For fair comparison and reduced overhead, we
adopt only the attention alignment and final distillation stages. In the original setup, attention
alignment was limited to 200M tokens (out of 3B total) due to the high cost of computing softmax
attention.

In our adaptation to the vision setting, we perform one full epoch of distillation using the
training set. The attention matrices from the Vim student are computed following the formulation
used in \textit{The Hidden Attention of Mamba Models} \citep{ali2024hidden}. In the second stage, we
apply KL divergence between the teacher and student logits as the distillation objective. To handle
depth mismatches between the teacher and student, we introduce a layer alignment function $g(l)$
that maps each student layer $l$ to its corresponding teacher layer. Hidden state matching is
omitted to isolate the effect of attention transfer and maintain efficiency.

\paragraph{Mamba in the Llama.} Mamba in the Llama \citep{wang2024mambainllama} introduces an
attention-initialized Mamba variant, using a linearized formulation as the initialization scheme for
distillation. Specifically, the linear projections of the Transformer teacher---including $Q$, $K$,
$V$, and the output projection---are mapped to the corresponding $C$, $B$, $X$, and output
projection layers in the Mamba student.

In our vision adaptation, we adopt the same strategy: the $Q$, $K$, $V$, and output projection
matrices from the ViT teacher are linearly mapped to the $C$, $B$, $X$, and output projection
modules in the Vim student. To address the mismatch in depth between teacher and student models, we
further introduce our layer alignment function $g(l)$ to assign supervision across corresponding
layers.

\paragraph{CAB(Ours).} We implement our bidirectional distillation framework tailored for visual
tasks. Specifically, we align both the forward and backward dynamic projections $B$, $C$ from the
student Vim with the $K$, $Q$ representations from the Transformer teacher (e.g., DeiT). This
alignment is performed layer-wise via a mapping function $g(l)$ and optimized jointly with soft KL
loss on logits. To improve compatibility between attention and state-space recurrence, we also
initialize the Mamba transition matrix $A \approx 0$ to simulate linear attention behavior. A full
description of the algorithm is provided in Algorithm~\ref{alg:attn_distill_vision}.

\begin{center}
\begin{minipage}{\columnwidth}
\footnotesize\raggedright
\refstepcounter{algocf}\label{alg:attn_distill_vision}
\hrule
\vspace{1mm}
\noindent\textbf{Algorithm~\thealgocf: Attention Distillation for Bidirectional Vision Mamba}
\vspace{1mm}
\hrule
\vspace{1mm}
\noindent\textbf{Input:} Teacher $\mathcal{T}$, bidirectional Mamba student $\mathcal{S}$, image tokens $x$, loss weight $\lambda$.
\par\noindent\textbf{Output:} Trained student model $\mathcal{S}$.
\vspace{1mm}
\begin{tabularx}{\linewidth}{@{}r@{\hspace{0.5em}}>{\raggedright\arraybackslash}X@{}}
1 & Initialize $A \approx 0$ such that $\bar{A} \approx I$; define layer mapping $g(l)$. \\
2 & For each batch $x \sim \mathcal{D}$, extract $\{K^{(l)},Q^{(l)}\}\leftarrow\mathcal{T}(x)$ and $\{B^{(l)}_{\rm fw},C^{(l)}_{\rm fw},B^{(l)}_{\rm bw},C^{(l)}_{\rm bw}\}\leftarrow\mathcal{S}(x)$. \\
3 & For each layer $l$, compute forward and backward bridge losses by matching $\phi_B(B)$ to $K^{(g(l))}$ and $\phi_C(C)$ to $Q^{(g(l))}$. \\
4 & Aggregate $\mathcal{L}_{\rm attn}=\frac{1}{L}\sum_l(\mathcal{L}^{(l)}_{\rm fw}+\mathcal{L}^{(l)}_{\rm bw})$ and compute $\mathcal{L}_{\rm KL}=\mathrm{KL}(p_T\|p_S)$. \\
5 & Update $\mathcal{S}$ with $\nabla(\mathcal{L}_{\rm attn}+\lambda\mathcal{L}_{\rm KL})$. \\
\end{tabularx}
\vspace{1mm}
\hrule
\end{minipage}
\end{center}

Table~\ref{tab:vision_training_recipe} summarizes the training configurations for visual
distillation experiments under different data regimes (1\%, 5\%, 10\%, 20\%). Models are trained on
ImageNet-1k using BF16 precision on a single NVIDIA A100 GPU.

\begin{table*}[tb]
\centering
\renewcommand{\arraystretch}{1.2}
\tablestyle{5pt}{1.15}
\caption{Training recipe for visual distillation experiments under different data regimes.}
\label{tab:vision_training_recipe}
\begin{adjustbox}{max width=\textwidth}
\begin{tabular}{lcccc}
\textbf{Config} & \textbf{1\% Data} & \textbf{5\% Data} & \textbf{10\% Data} & \textbf{20\% Data} \\
\shline
Optimizer           & AdamW     & AdamW     & AdamW     & AdamW \\
Learning rate       & 5e-4      & 5e-4      & 5e-4      & 5e-4 \\
Weight decay        & 0.05      & 0.05      & 0.05      & 0.05 \\
Training epochs     & 3000      & 600       & 300       & 150 \\
Optimizer momentum  & (0.9, 0.999) & (0.9, 0.999) & (0.9, 0.999) & (0.9, 0.999) \\
Batch size          & 64        & 64        & 64        & 64 \\
LR schedule         & Cosine decay & Cosine decay & Cosine decay & Cosine decay \\
\end{tabular}
\end{adjustbox}
\end{table*}

\begin{table*}[tb]
\centering
\renewcommand{\arraystretch}{1.2}
\tablestyle{5pt}{1.15}
\caption{Training recipe for LLM distillation experiments at different scales.}
\label{tab:llm_training_recipe}
\begin{adjustbox}{max width=\textwidth}
\begin{tabular}{lcc}
\textbf{Config} & \textbf{100M-scale} & \textbf{1B-scale} \\
\shline
Dataset                & OpenWebText & C4 corpus \citep{raffel2020exploring} \\
Max sequence length    & 1024 & 2048 \\
Batch size (per device)& 4 & 32 \\
Learning rate          & 2e-5 & 5e-5 \\
Training tokens        & 200M (Stage 1) / 2B or 4B (Stage 2) & 200M (Stage 1) / 2.8B (Stage 2) \\
Optimizer              & \multicolumn{2}{c}{AdamW} \\
Adam $\beta$           & \multicolumn{2}{c}{(0.9, 0.999)} \\
Adam $\epsilon$        & \multicolumn{2}{c}{1e-8} \\
Precision              & \multicolumn{2}{c}{bfloat16 (BF16)} \\
Loss functions         & \multicolumn{2}{c}{Attention alignment (Stage 1) + KL divergence (Stage 2)} \\
\end{tabular}
\end{adjustbox}
\end{table*}

\subsection{Implementation Details for LLM Distillation Methods}
\label{app:llm-implementation}
For the language modeling setting, we adapt both the \textbf{Transformers to SSMs (MOHAWK)} and
\textbf{Mamba in the Llama} baselines into our distillation pipeline using the Phi-Mamba
architecture as the student.

\paragraph{Transformers to SSMs (MOHAWK).} For MOHAWK, we follow its original two-stage design by
performing attention matrix alignment followed by output-level knowledge distillation. Specifically,
we distill attention representations using 200M tokens in the first stage, and continue soft KL
divergence-based distillation over 2B or 4B tokens depending on the experimental setup. To ensure
fairness and training feasibility, we omit the intermediate hidden state alignment stage. Layer
mismatches between the Transformer teacher (DistilGPT2) and the student are handled using a layer
alignment function $g(l)$.

\paragraph{Mamba in the Llama.} For Mamba in the Llama, we replicate their projection initialization
strategy by directly mapping the Transformer's $Q$, $K$, $V$, and output projection weights to the
Mamba's $C$, $B$, input, and output projection layers, respectively. We then train the student using
the same two-stage strategy on 2B or 4B tokens without additional supervision.

\paragraph{CAB(Ours).} We apply our attention-level distillation framework to the LLM setting. The
student model (Phi-Mamba) is trained in two stages: first aligning the token-wise $B$, $C$
projections to the teacher's $K$, $Q$ via layer-mapped MLPs, followed by KL-based soft distillation
on teacher logits. Detailed pseudocode for this training process is provided in
Algorithm~\ref{alg:nlp_distillation}.

\begin{center}
\begin{minipage}{\columnwidth}
\footnotesize\raggedright
\refstepcounter{algocf}\label{alg:nlp_distillation}
\hrule
\vspace{1mm}
\noindent\textbf{Algorithm~\thealgocf: Two-Stage Attention Distillation for Mamba in Language Modeling}
\vspace{1mm}
\hrule
\vspace{1mm}
\noindent\textbf{Input:} Teacher $\mathcal{T}$, causal Mamba student $\mathcal{S}$, tokenized dataset $\mathcal{D}$.
\par\noindent\textbf{Output:} Trained student model $\mathcal{S}$.
\vspace{1mm}
\begin{tabularx}{\linewidth}{@{}r@{\hspace{0.5em}}>{\raggedright\arraybackslash}X@{}}
1 & \textbf{Stage 1:} for the first $N_1$ tokens, extract $\{K^{(l)},Q^{(l)}\}\leftarrow\mathcal{T}(x)$ and $\{B^{(l)},C^{(l)}\}\leftarrow\mathcal{S}(x)$. \\
2 & For each layer $l$, compute $\mathcal{L}^{(l)}_{\rm attn}=\|\phi_B(B^{(l)})-K^{(g(l))}\|^2+\|\phi_C(C^{(l)})-Q^{(g(l))}\|^2$. \\
3 & Update $\mathcal{S}$ using $\nabla \frac{1}{L}\sum_{l=1}^{L}\mathcal{L}^{(l)}_{\rm attn}$. \\
4 & \textbf{Stage 2:} continue to $N_2$ tokens with $\mathcal{L}_{\rm KL}=\mathrm{KL}(p_T\|p_S)$. \\
5 & Update $\mathcal{S}$ using $\nabla\mathcal{L}_{\rm KL}$. \\
\end{tabularx}
\vspace{1mm}
\hrule
\end{minipage}
\end{center}






\section{Additional Visualizations}
We provide supplementary training trajectories and qualitative attention heatmaps to illustrate the
optimization behavior and visual responses of CAB under limited-supervision ImageNet training.



\makeatletter
\setlength{\@fptop}{0pt}
\setlength{\@dblfptop}{0pt}
\makeatother
\begin{figure*}[!tp]
  \centering
  \begin{minipage}[t]{0.45\linewidth}
    \centering
    \includegraphics[width=\linewidth]{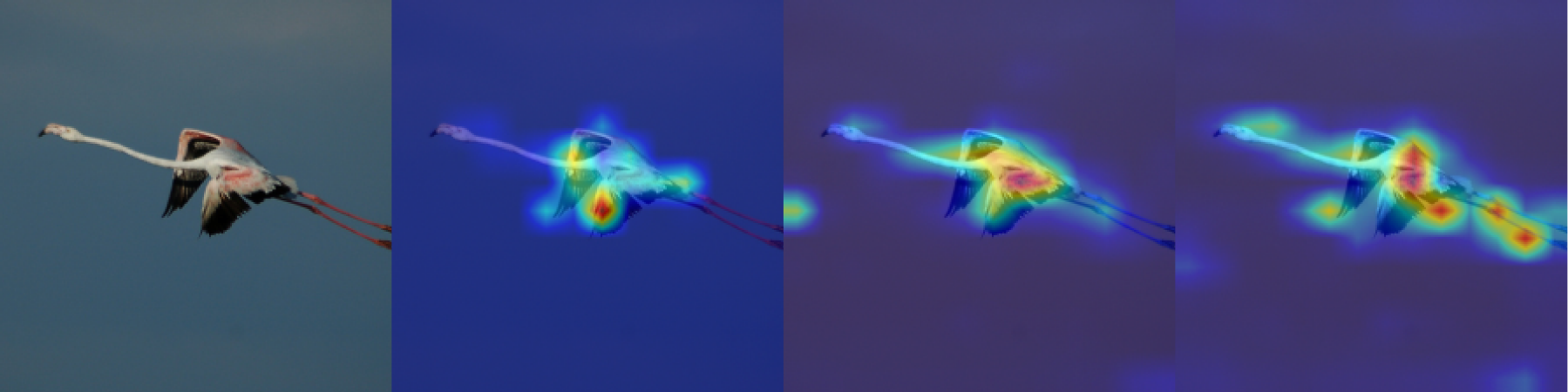}
  \end{minipage}
  \begin{minipage}[t]{0.45\linewidth}
    \centering
    \includegraphics[width=\linewidth]{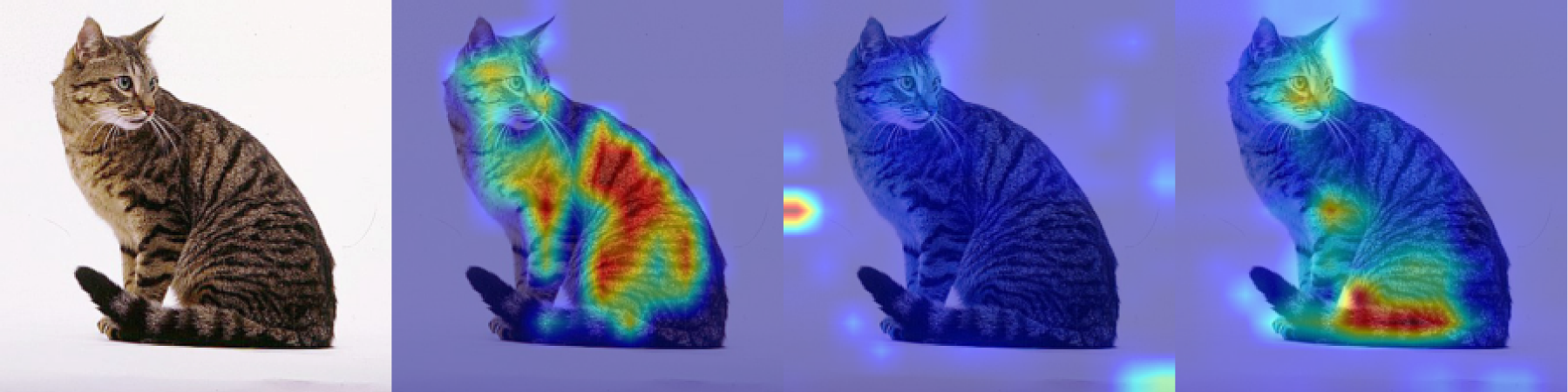}
  \end{minipage}
  
  \begin{minipage}[t]{0.45\linewidth}
    \centering
    \includegraphics[width=\linewidth]{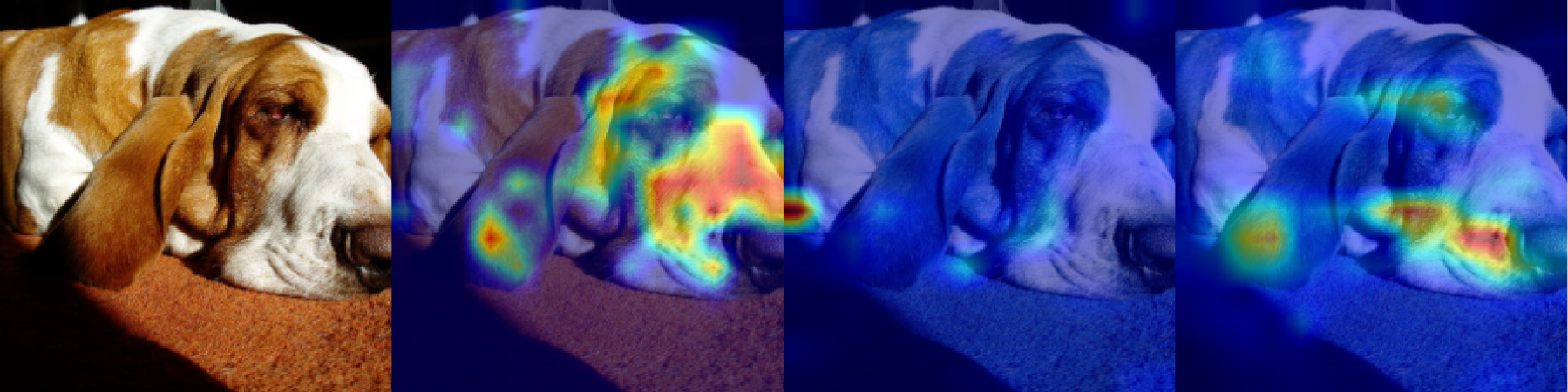}
  \end{minipage}
  \begin{minipage}[t]{0.45\linewidth}
    \centering
    \includegraphics[width=\linewidth]{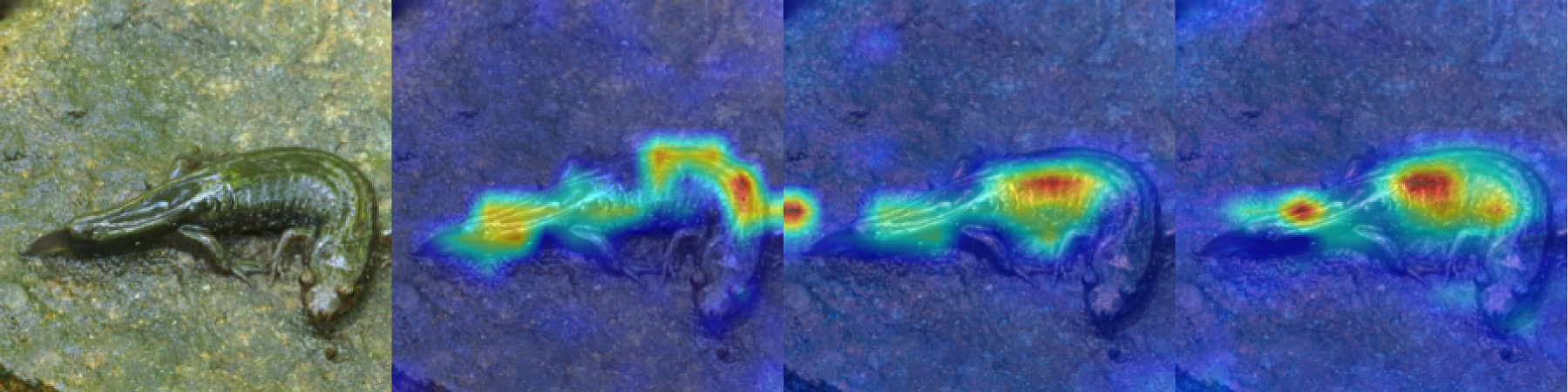}
  \end{minipage}
  
  \begin{minipage}[t]{0.45\linewidth}
    \centering
    \includegraphics[width=\linewidth]{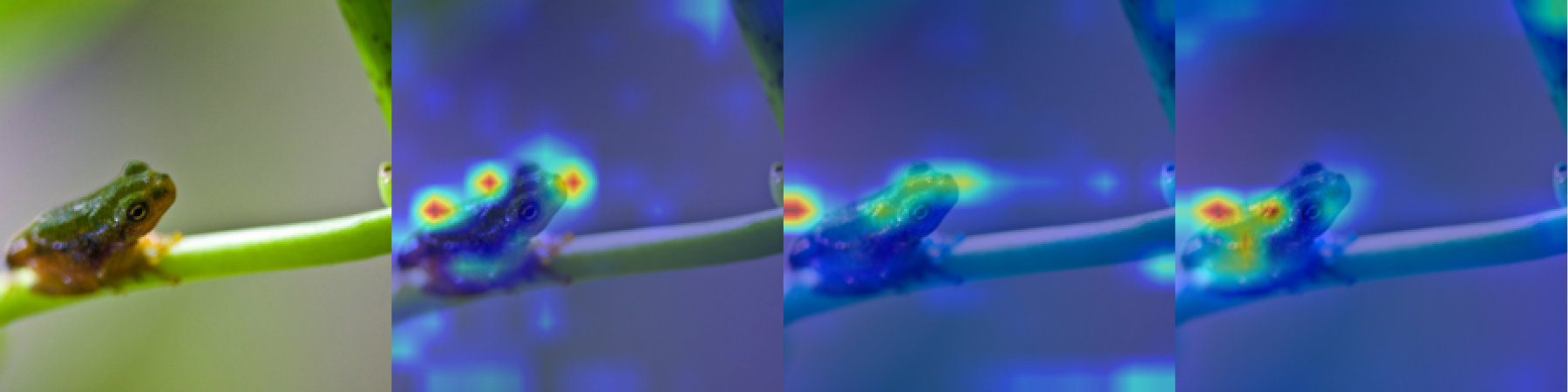}
  \end{minipage}
  \begin{minipage}[t]{0.45\linewidth}
    \centering
    \includegraphics[width=\linewidth]{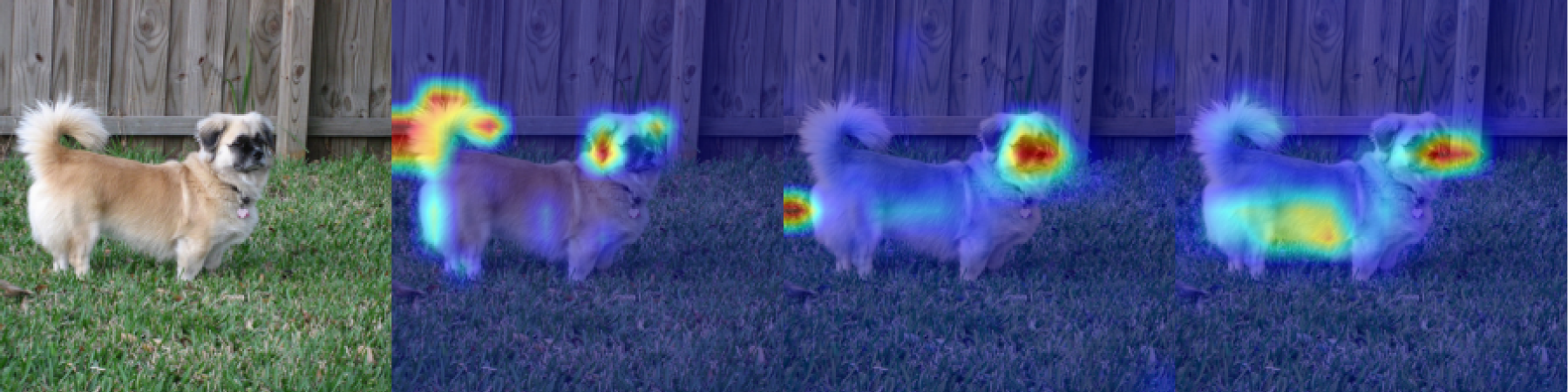}
  \end{minipage}
  
  \begin{minipage}[t]{0.45\linewidth}
    \centering
    \includegraphics[width=\linewidth]{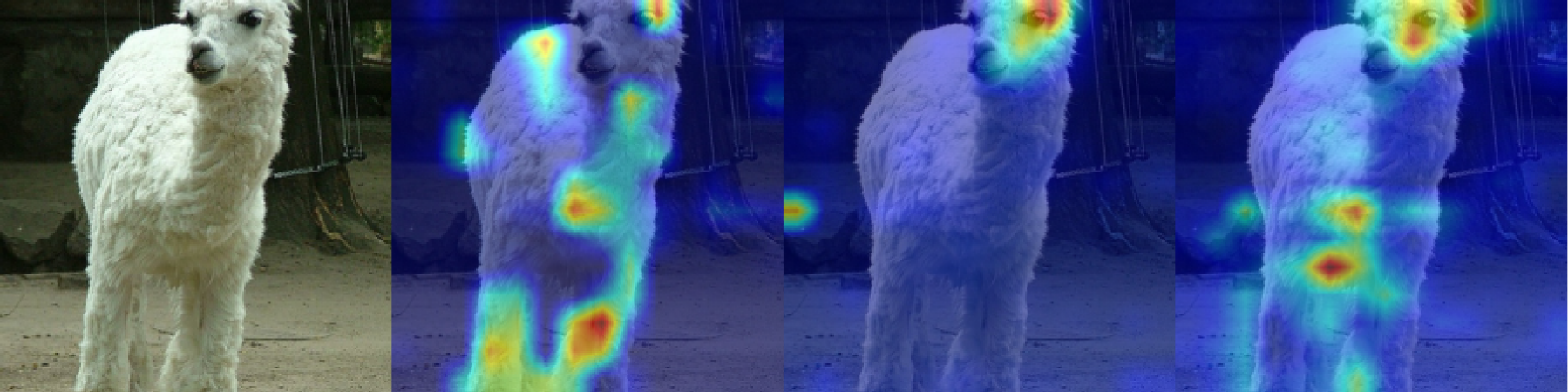}
  \end{minipage}
  \begin{minipage}[t]{0.45\linewidth}
    \centering
    \includegraphics[width=\linewidth]{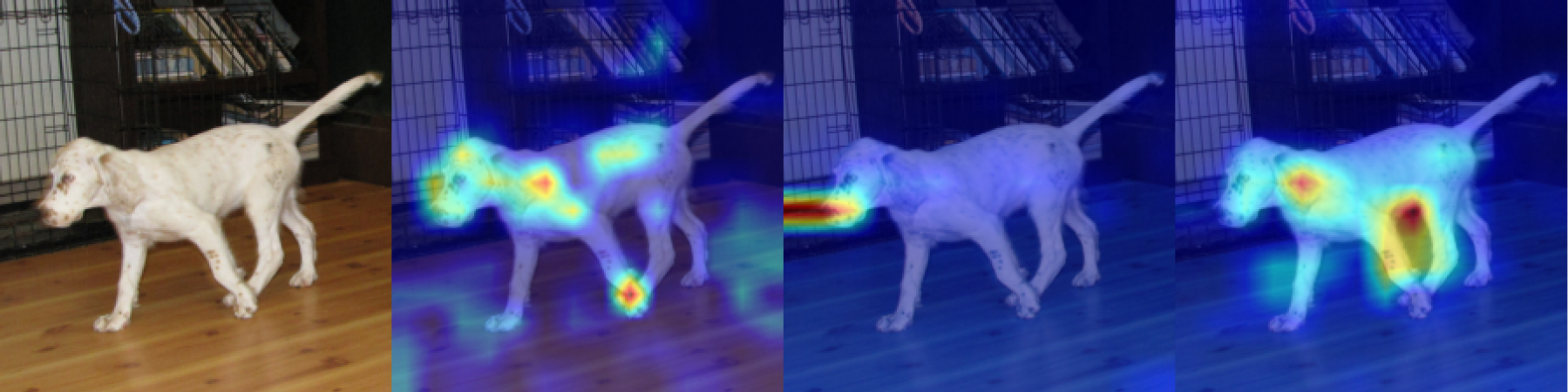}
  \end{minipage}
  
  \caption{
    \textbf{Visualization of attention heatmaps across teacher and student models.}
    In each group of images, we show the original image followed by heatmaps generated from:
    (1) the pretrained ViT teacher, 
    (2) a Vim model trained on the full ImageNet training set, and 
    (3) our Vim student trained with CAB using only 10\% of the full ImageNet training set.
    Despite limited supervision, our method produces more concentrated and semantically meaningful attention responses.
  }
  
  \label{fig:six_vis}
\end{figure*}

\end{document}